\documentclass{article}

\usepackage{arxiv}
\usepackage[T1]{fontenc}    
\usepackage{hyperref}       
\usepackage{url}            
\usepackage{booktabs}       
\usepackage{amsfonts}       
\usepackage{nicefrac}       
\usepackage{microtype}      
\usepackage{lipsum}
\usepackage{graphicx}
\usepackage{amsmath}
\usepackage{algorithm}
\usepackage{algorithmic}
\graphicspath{ {./images/} }

\title{Robust face recognition based on the wing loss and the $\ell_1$ regularization}

\author{
	Yaoyao Yun  \\
	College of Mathematics and Statistics\\
	Chongqing University\\
	Chongqing, China \\
	\texttt{202206021026@stu.cqu.edu.cn} \\
	\And
	Jianwen Xu \\
	National Elite Institute of Engineering\\
	Chongqing University\\
	Chongqing, China  \\
	\texttt{xjw@cqu.edu.cn} \\
}

\begin{document}
	\maketitle
	\begin{abstract}
		In recent years, sparse sampling techniques based on regression analysis have witnessed extensive applications in face recognition research. Presently, numerous sparse sampling models based on regression analysis have been explored by various researchers. Nevertheless, the recognition rates of the majority of these models would be significantly decreased when confronted with highly occluded and highly damaged face images. In this paper, a new wing-constrained sparse coding model(WCSC) and its weighted version(WWCSC) are introduced, so as to deal with the face recognition problem in complex circumstances, where the alternating direction method of multipliers (ADMM) algorithm is employed to solve the corresponding minimization problems. In addition, performances of the proposed method are examined based on the four well-known facial databases, namely the ORL facial database, the Yale facial database, the AR facial database and the FERET facial database. Also, compared to the other methods in the literatures, the WWCSC has a very high recognition rate even in complex situations where face images have high occlusion or high damage, which illustrates the robustness of the WWCSC method in facial recognition.
	\end{abstract}
	
	\keywords{sparse sampling; weight learning; face recognition; robustness} 

	\section{Introduction}
	Face recognition is a branch of visual pattern recognition. Humans usually understand the world through visual pattern recognition, because humans get visual information through the naked eye. However, both photos and videos are recognized by computers as matrices, and the elements of the matrices are individual pixels. In contrast to human perception, computers process images by analyzing these matrices, which is a fundamental difference in how recognition tasks are approached. The challenge in face recognition lies in the computer's ability to identify and distinguish between different faces, despite variations in lighting, poses and expressions. This is achieved through the use of algorithms that can extract features from the pixel matrices and compare them to a database of known faces. The development of such algorithms has been a significant area of research in the field of computer vision and machine learning.
	
	Facial recognition technology, which achieves the identification of an individual's identity by comparing the digital image of a human face with the known face database. In fact, face recognition technology has experienced many stages in its development, such as the initial template-matching based approach, the machine learning-based approach \cite{ref-journal1}, and the deep learning approach \cite{ref-journal2} nowadays. Early face recognition methods primarily relied on template-based matching to identify individuals. This approach achieved face recognition by comparing the similarity between a given face image and a predefined template. However, this method is sensitive to the interference of illumination, expressions and some other factors, and it is difficult to achieve accurate face recognition \cite{ref-journal1}. With the development of machine learning technology, face recognition algorithms based on machine learning have received much attention, such as the Eigenface \cite{ref-journal3} and the Fisherface \cite{ref-journal4}. These algorithms extract the features of face images and use machine learning algorithms for classification and recognition \cite{ref-journal4}. Compared to template matching methods, machine learning-based face recognition algorithms seem to achieve higher accuracy and be more robust. Furthermore, deep learning technology can extract and classify high-dimensional features of face images by training deep neural network models \cite{ref-journal5}, such as the FaceNet \cite{ref-book1}, VGGFace \cite{ref-book2}, etc.
	
	However, with the progress and development of technology, sparse sampling method has been widely used in the field of face recognition. In this method, it is assumed that the signal can be represented as a linear combination of a sparse coefficient vector and an atomic dictionary, where the atomic dictionary is a set of sample points that can represent various parts of the signal. Therefore, the aim of this method is to extract the sparse coefficient vector based on as few sampling data as possible. After that, the signal can be reconstructed by the sparse coefficient vector and the atomic dictionary. Therefore, for a signal $y \in R^m$, it can be expressed as
	
	\begin{equation}
		y=Ax+\varepsilon,
	\end{equation}
	
	where $\varepsilon \in R^m$ denotes the measurement noise, $A$ is the atomic dictionary and $x$ is the sparse coefficient vector.
	
	In 2009, Wright et al. proposed the theory of compressed perception for face recognition \cite{ref-journal6}. The key point of the sparse sampling problem is how to reconstruct sparse signal. Given the sparse nature of the signal, the reconstruction process involves identifying the non-zero elements of the sparse coefficient vector $x$. This is typically achieved through optimization techniques that aim to find the sparsest solution that satisfies the equation $y=Ax$. Therefore, the reconstruction algorithm must effectively exploit the structure of the atomic dictionary $A$ and the sparsity of the signal to recover the original signal from the noisy measurements. Sparsity means that most of the elements in the signal vector $x$ are assumed to be zero, which leads to the following optimization problem:
	
	\begin{equation}
		\label{l0}
		\min \Vert x \Vert_0 \qquad s.t.\Vert Ax-y \Vert\leq\epsilon,
	\end{equation}
	
	where $\Vert x \Vert_0$ denotes the $\ell_0$-norm, counting the non-zero entries of $x$ and $ \epsilon\geq 0 $ denotes the level of sampling noise. However, the above optimization problems is a NP-hard problem. To reconstruct the sparse signal, one alternative widely used approach is to replace $\ell_0$-norm with $\ell_1$-norm. Therefore, this optimization problem (\ref{l0}) can be rewritten as:
	
	\begin{equation}
		\min \Vert x \Vert_1 \qquad s.t.\Vert Ax-y \Vert\leq\epsilon,
	\end{equation}
	
	where $\Vert x \Vert_1=\sum_{i=1}^{n}x_i$ denoting the $\ell_1$-norm is a concave approximation of the $\ell_0$-norm. This approximation allows for the relaxation of the original NP-hard problem into a tractable convex optimization problem, which can be efficiently solved using various algorithms such as linear programming or iterative thresholding methods. The $\ell_1$-norm minimization has been shown to be effective in recovering sparse signals in the presence of noise, which is particularly useful in signal processing and compressed sensing applications. Finally, it is obvious that we only need to solve the following unconstrained minimization problem:
	
	\begin{equation}
		\min \{\frac{1}{2}\Vert Ax-y \Vert_2^2+\lambda\Vert x \Vert_1\}.
	\end{equation}

	Various algorithms have been proposed in the literatures to deal with this well-known LASSO problem, such as the alternating direction method of multipliers (ADMM) \cite{ref-journal7}, Bregman methods \cite{ref-journal8}, the Frank–Wolfe algorithm \cite{ref-journal9} and the iterative thresholding methods \cite{ref-journal10}. When the measurement matrix $A$ satisfies the restricted isometry property (RIP) with a sufficient high order, the sparse solution can be obtained through these methods. Therefore, the $\ell_1$-norm regularization method is particularly useful in scenarios where the signal of interest is sparse. In the meantime, this method has also been widely used in some other situations, including signal processing, image reconstruction and machine learning, where the goal is to extract meaningful information from noisy and incomplete data \cite{ref-journal11}.
	
	However, most of the methods mentioned above cannot achieve efficient recognition. On the one hand, this failure is due to the similarity between faces and the variability of the faces. Images of faces from different individuals have similar biometric characteristics. This similarity makes it difficult to identify a face image. Also, for different postures, illuminations, angles and expressions, the face images are unstable, which would lead to the variability. On the other hand, the face images may be partly blocked or damaged, which is not conducive to face recognition. Similarly, there may exist noises in face images, which can also cause difficulties in face recognition. Therefore, it is necessary to study robust face recognition methods \cite{ref-journal12}.
	
	In recent years, many robust face recognition methods are proposed in the literatures. Zhang et al. (2011) explained the possibility of sparse representation in the case of sufficient samples and verified the advantages of sparse representation for face recognition. They introduced a collaborative representation classifier (CRC) based on $\ell_2$-norm constraints and further proposed its robust version (RCRC) \cite{ref-book3}. Zhou et al. (2015) considered a classification method for face recognition based on $\ell_\frac{1}{2}$ regularization, which balanced the sparse representation classifier (SRC) and the CRC through an iterative Tikhonov regularization (ITR) \cite{ref-journal13}. Zhang et al. (2017) solved the robust face recognition problem via the iterative re-constrained group sparse classifier (IRGSC) with adaptive weights learning \cite{ref-journal14}. Lei et al. (2020) proposed the weighted Huber constrained sparse coding (WHCSC) and established a robust weighted regression model with sparse constraints for face recognition \cite{ref-journal12}.
	
	In this paper, a new robust face recognition method based on the weighted wing loss constrained sparse coding (WWCSC) is proposed. Performances of the proposed method are examined by various experiments, and it is demonstrated to outperform some other robust face recognition methods in the literatures, especially when face images are partly blocked or damaged.

	\section{The recognition method based on sparse robust coding}
	
	In the face recognition problem, each gray image of a face can be represented as a gray matrix. For ease of representation, we can stack the gray matrix as a column vector. It is assumed that there are $k$ classes of face images and $k$ is known. Let $ A_i=\left[ v_{i,1},v_{i,2},\cdots,v_{i,n_i}\right]\in R^{m\times n_i}$, which denotes training samples of class-$i$, where $n_i$ is the number of training samples of class-$i$ and each column of $A_i$ represents a face image from class-$i$. Therefore, if given enough training samples from class-$i$, any new test sample from the same class can be represented linearly by the columns of the training samples $A_i$, i.e. 
	\begin{equation}
		x_{i,1}v_{i,1}+x_{i,2}v_{i,2}+\cdots+x_{i,n_i}v_{i,n_i},
	\end{equation}
	where $x_{i,1},x_{i,2},\cdots,x_{i,n_i} $ are the corresponding coefficients.
	
	Furthermore, let $A=\left[ A_1,A_2,\cdots,A_k\right]\in R^{m\times n}$, where $ n=\sum_{i=1}^{k} n_i$ is the total number of training samples and assume $k$ is very large, then for a given new test sample $ y \in R^{m\times 1}$, it can be linearly represented as:
	\begin{equation}
		y=Ax_0,
	\end{equation}
	where the vector $x_0 $ must be sparse since the test image only belongs to a certain class. To get the sparse vector, we need to consider the following sparse optimization problem:
	\begin{equation}
		\label{l2-loss}
		\min_{x} \{\Vert Ax-y \Vert_2^2+\lambda\Vert \alpha \Vert_1\} \qquad s.t.\alpha=x,
	\end{equation}
	where $ \lambda$ is the penalty coefficient for the $\ell_1$-norm. The essence of formula (\ref{l2-loss}) is the sparse constraint when the residuals of least squares estimation obey the Gaussian distribution. However, when residuals obey the Laplacian distribution, the sparse coding problem is:
	\begin{equation}
		\label{l1-loss}
		\min_{x}  \{\Vert Ax-y \Vert_1+\lambda\Vert \alpha \Vert_1\} \qquad s.t.\alpha=x.
	\end{equation}
	
	Sparse sampling is a technique which can capture high-level correlated structures in images and represent signals with as few atoms as possible in a given over-complete dictionary. In a word, sparse sampling attempts to construct or approximately represent complex signals or images using the minimum number of basic elements (atoms in the dictionary), while retaining important features and structural information in the image. However, the problem is whether the fidelity term ($\Vert Ax-y \Vert_1 $ or $ \Vert Ax-y \Vert_2$) is sufficiently effective to describe the fidelity of the signal, especially when the signal has noise or abnormal values \cite{ref-journal12}. Here, fidelity generally refers to the degree of similarity between the original signal and the signal which has been processed or encoded. For formula (\ref{l2-loss}) and formula (\ref{l1-loss}), the $\ell_2$ norm (Euclidean distance) or $\ell_1$ norm (Manhattan distance) are employed to define fidelity. This definition is based on the Maximum A Posteriori Probability (MAP) and assumes that the residuals after encoding (i.e., the difference between the original signal and the encoded signal) follow a Gaussian distribution (normal distribution) or a Laplace distribution. However, in real practice, the distribution of residuals is unknown, and it may not be a good choice to follow a certain distribution of a single hypothetical residual, especially when occlusion, camouflage or corruption occurs in facial images. Therefore, a fidelity item that uses a single norm in a sparse coding model may not be robust in these cases\cite{ref-journal12}.
	
	To solve this problem, we use the wing loss to replace the $\ell_1$-norm loss or the $\ell_2$-norm loss. This type of loss was proposed by Feng et al. \cite{ref-book4}, which can be defined as:
	\begin{equation}
		wing(x)=\begin{cases}
			\omega\ln(1+\lvert x \rvert)/\epsilon,   & if \;\lvert x \rvert<\omega, \\
			x-c,  & otherwise.
		\end{cases}
	\end{equation}
	where the non-negative parameter $\omega$ limits the value range of the nonlinear part to (-$\omega$,$\omega$), the parameter $\epsilon$ limits the curvature of the nonlinear region and $C=\omega-\omega ln(1+\omega/\epsilon)$ is a constant that smoothly connects the linear and nonlinear parts defined in the piecewise function\cite{ref-book4}.  Based on the theory by Byod \cite{ref-book5} and the ADMM algorithm, we firstly introduced the following wing constrained sparse coding model (WCSC):
	\begin{equation}
		\min_{x} g(z)+\lambda\Vert \alpha \Vert_1 \qquad s.t.z=Ax-y,\alpha=x,
	\end{equation}
	where
	\begin{equation}
		g(z)=\left\{
		\begin{aligned}
			\omega\ln(1+\Vert z \Vert_1)/\epsilon & , & if \;\Vert z \Vert_1<\omega, \\
			\Vert z \Vert_1-C & , & otherwise.
		\end{aligned}
		\right.\quad
	\end{equation}
	Under specific circumstances, as the value of $\lambda$ increases, the sparsity of $x$ becomes more pronounced.
	
	In real life, the sample data may contain outliers. To further diminish the impact of noise or outliers in the training sample, an effective weighted approach is to assign lower weights to the outliers. In robust sparse representation-based classifier (RSRC), the corresponding minimization problem could be converted into an iteratively reweighted sparse coding problem \cite{ref-journal15}
	
	Based on the weight vector in RSRC and the wing constrained sparse coding model (WCSC) mentioned above, we also consider a weighted wing constrained sparse coding model (WWCSC) in this paper. Actually, impacts of the noise and outliers can be effectively mitigated through the utilization of the weight vector and the WCSC model. Subsequently, the $\ell_1$-norm minimization can be addressed by  the ADMM algorithm. Numerous data experiments conducted in some open face databases demonstrate that the WWCSC model exhibits an excellent classification effect, particularly when confronted with complex facial images such as occlusion, corrosion, and so on.
	
	The WWCSC model can be expressed as
		\begin{equation}
			\label{wwcsc}
			\min_{x}  g(z)+\lambda\Vert \alpha \Vert_1 \qquad s.t.z=w^T(Ax-y),\alpha=x,
		\end{equation}

	where $w=(w_1,w_2,\cdots,w_m) \in R^{m\times1}$ is the weight vector. In particular, the weight of the $i$-th sample $w_i$ is defined by the following sigmoid function:

		\begin{equation}
			\label{weight1}
			w_i(e_i)=\frac{1}{1+exp(-q(\frac{\delta-e_i^2}{\delta}))},
		\end{equation}

	where $e_i$ is the residual and $\delta$ is the residual threshold. Obviously, $\delta-e_i^2$ represents the distance between the residual and the threshold. Furthermore, the parameter $q$ influences the penalty rate of the weight. Thus, the sigmoid function constrains the weight values within the range of $[0,1]$. Also, when the residual is greater than $\delta$, the weight is less than 0.5, while if the residual is less than $\delta$, the weight is greater than 0.5. Let $\Psi=(e_1^2,e_2^2,\cdots,e_m^2)$, and then rearrange $\Psi$ to $\Psi_a$ in a descending order ranging from the smallest to the largest. In addition, if we assume $k=\lfloor\tau m\rfloor$, where $\tau \in [0,1]$ and $\lfloor\tau m\rfloor$ represents the largest integer less than $ \tau m$, then the parameter $\delta$ can be expressed as $\delta=\Psi_a(k)$ following the article \cite{ref-journal15}. For the sake of facilitating the calculation, formula (\ref{weight1}) can be rewritten as

		\begin{equation}
			w_i(e_i)=\frac{exp(-\mu e_i^2+\mu \delta)}{1+exp(-\mu e_i^2+\mu \delta)},
		\end{equation}

	where the parameter $\mu=\frac{q}{\delta}$.
	
	For the optimization problem (\ref{wwcsc}), we can use the ADMM algorithm to solve it. The Lagrange function corresponding to the above optimization problem is 
	\begin{equation}
		L(x,z,\alpha,h_1,h_2)=g(z)+\lambda \Vert\alpha\Vert_1+\langle h_1,w^T(Ax-y)-z\rangle+\langle h_2,\alpha-x\rangle,
	\end{equation}
	where $h_1$ and $h_2$ are Lagrange multipliers and $\langle . \rangle$ denotes the inner product.
	
	In the ADMM algorithm, while one variable is updated, the others are assumed to be fixed as constants, which gradually approximates the optimal solution by minimizing the Lagrange function. This method has been widely applied in statistical learning and machine learning due to its effective handling of convex optimization problems with equality constraints, fast processing speed, and good convergence. Furthermore, by adding a quadratic penalty term to the original Lagrange function, the improved augmented Lagrange function is defined as
	\begin{equation}
		\begin{split}
			L_{\rho_1,\rho_2}(x,z,\alpha,h_1,h_2)=g(z)+\lambda \Vert\alpha\Vert_1+\langle h_1,w^T(Ax-y)-z\rangle+\langle h_2,\alpha-x\rangle \\ 
			+\frac{\rho_1}{2}\Vert w^T(Ax-y)-z\Vert_2^2+\frac{\rho_2}{2}\Vert \alpha-x\Vert_2^2, 
		\end{split}
	\end{equation}
	where $\rho_1>0 \,,\;\rho_2>0$. In fact, the augmented Lagrange function could also be rewritten as
	\begin{equation}
		\begin{split}
			L_{\rho_1,\rho_2}(x,z,\alpha,h_1,h_2)=g(z)+\lambda \Vert\alpha\Vert_1+\frac{\rho_1}{2}\Vert w^T(Ax-y)-z+u_1\Vert_2^2\\+\frac{\rho_2}{2}\Vert \alpha-x+u_2\Vert_2^2,   
		\end{split}
	\end{equation}
	where $u_1=\frac{h_1}{\rho_1}$ , $u_2=\frac{h_2}{\rho_2}$. By the well-known ADMM algorithm, the above augmented Lagrange function could be iterated as follows:
	\begin{align}
		x^{(k+1)}&= arg\;min\,\frac{\rho_1}{2}\Vert W^T(Ax^{(k)}-y)-z^{(k)}+u_1^{(k)}\Vert_2^2+\frac{\rho_2}{2}\Vert \alpha^{(k)}-x^{(k)}+u_2^{(k)}\Vert_2^2 \\
		z^{(k+1)}&= arg\;min\,g(z^{(k)})+\frac{\rho_1}{2}\Vert W^T(Ax^{(k+1)}-y)-z^{(k)}+u_1^{(k)}\Vert_2^2   \\
		\alpha^{(k+1)}&= arg\;min\,\lambda \Vert \alpha^{(k)}\Vert_1+\frac{\rho_2}{2}\Vert \alpha^{(k)}-x^{(k+1)}+u_2^{(k)}\Vert_2^2    \\
		u_1^{(k+1)}&=u_1^{(k)}+W^T(Ax^{(k+1)}-y)-z^{(k+1)}\\
		u_2^{(k+1)}&=u_2^{(k)}+\alpha^{(k+1)}-x^{(k+1)}
	\end{align}
	where $W=diag(w)=diag(w_1,w_2,\cdots,w_m)$. Subsequently, the aforementioned sub-optimization problems can be addressed individually:
	\begin{align}
		\label{x}
		x^{(k+1)}&= [\rho_1A^TWW^TA+\rho_2I]^{-1}[\rho_1A^TW(W^Ty+z^{(k)}-u_1^{(k)})+\rho_2(\alpha^{(k)}+u_2^{(k)})] \\
		\label{z}
		z^{(k+1)}&=\left\{\begin{aligned}
			S_{\frac{\eta}{\rho_1{(\varepsilon+\Vert z^{(k)} \Vert_1)}}}(W^T(Ax^{(k+1)}-y)+u_1^{(k)}) & , & \Vert z^{(k)} \Vert_1<\eta, \\
			S_{\frac{1}{\rho_1}}(W^T(Ax^{(k+1)}-y)+u_1^{(k)}) & , & \Vert z^{(k)} \Vert_1 \geq \eta.
		\end{aligned}
		\right.\quad
		\\
		\label{alpha}
		\alpha^{(k+1)}&= S_{\frac{\lambda}{\rho_2}}(x^{(k+1)}-u_2^{(k)})    \\
		\label{u_1}
		u_1^{(k+1)}&=u_1^{(k)}+W^T(Ax^{(k+1)}-y)-z^{(k+1)}\\
		\label{u_2}
		u_2^{(k+1)}&=u_2^{(k)}+\alpha^{(k+1)}-x^{(k+1)}
	\end{align}
	where the S operator is defined as

		\begin{equation}
			S_k(a)=\left\{
			\begin{aligned}
				a-k & , &  a>k \\
				0 & , & \vert a \vert \leq k\\
				a+k & , & a<-k
			\end{aligned}
			\right.\quad.
		\end{equation}

	For a given test sample $y\in R^{m\times1}$, which is assumed to belong to one category in the training set, the sparse representation $\hat{x}$ can be calculated by the following Algorithm \ref{alg:1}. Ideally, the non-zero terms of the estimator are associated only with a certain category in the training set. In such cases, it is quite easy to determine the category to which the test sample belongs. However, in practical applications, noise or modeling errors may result in numerous non-zero terms in the obtained estimator, and these non-zero terms are associated with multiple categories in the training set. For such situations, numerous possible classifiers are designed to address this issue. For example, we can pick out the largest one among the estimator and attribute the test sample $y$ to the category associated with it. However, the aforementioned method does not take into account the utilization of the subspace structure associated with the images in face recognition. As can be seen from the previous description, there exists a linear structure among the images in the model. To better utilize this linear structure, the test samples are reconstructed using the training samples of each category at first, and then classification is carried out based on the differences between the reconstructed samples and the test samples.
	
	\begin{algorithm}
		\renewcommand{\algorithmicrequire}{\textbf{Input:}}
		\renewcommand{\algorithmicensure}{\textbf{Output:}}
		\caption{Weighted wing constrained sparse coding model}
		\label{alg:1}
		\begin{algorithmic}[1]
			\REQUIRE The atomic dictionary $A$,  test sample $y$
			\ENSURE The estimate $\hat{x}$
			\STATE Given the atomic dictionary $A$ and test sample $y$, select the appropriate parameter $\lambda,q,\tau,\rho_1,\rho_2,N_{ither},\tau_0$
			\STATE Initialize $x^{(0)},z^{(0)},\alpha^{(0)},u_1^{(0)},u_2^{(0)}$
			\STATE Calculate the weight of the initialization
			\FORALL{$k=0,1,\cdots,N_{ither}$}
			\STATE Update $x$ based on Equation~\ref{x}
			\STATE Update $z$ based on Equation~\ref{z}
			\STATE Update $\alpha$ based on Equation~\ref{alpha}
			\STATE Update $u_1$ based on Equation~\ref{u_1}
			\STATE Update $u_2$ based on Equation~\ref{u_2}
			\STATE Update the weight $w$ based on Equation $$w_i^{(k+1)}(e_i^{(k+1)})=\frac{exp(-\mu (e_i^{(k+1)})^2+\mu \delta)}{1+exp(-\mu (e_i^{(k+1)})^2+\mu \delta)}$$
			\ENDFOR
			\STATE Criteria for exiting a loop:$$\tau^{(k+1)}=\Vert x^{(k+1)}-x^{(k)}\Vert_2/\Vert x^{(k)}\Vert_2$$ If $\tau^{(k+1)}<\tau_0$ , then exit loop
			\STATE \textbf{return} The estimator $\hat{x}$
		\end{algorithmic}  
	\end{algorithm}
	
	For each class $i$, suppose $\delta_i:R^n\rightarrow R^n $ is the characteristic function that selects the coefficients related to the $i$-th class. Then, for given $x\in R^n$, non-zero elements of $\delta_i(x)$ are just the items in $x$ associated with category $i$. Based on the characteristic functions, we can reconstruct the test sample by using the coefficients associated with all the training samples of the $i$-th category. Therefore, the given test sample $y$ can be estimated as $\hat{y_i}=A\delta_i(\hat{x})$. Then, the recognition of $y$ can be realized based on these approximations by assigning it to the object class that minimizes the residual between $y$ and $\hat{y_i}$:

		\begin{equation}
			\min\;r_i(y)\doteq\Vert y-A\delta_i(\hat{x}) \Vert_2.
		\end{equation}

	The computational costs of most classification algorithms are associated with the dimensions of the input samples. In numerous scenarios of practical applications, the dimensions of the data might be extremely high, particularly in issues related to image classification. Thus, it is quite meaningful to reduce the dimensions of the data. Various dimensional reduction methods have been proposed in the literatures, such as the principal component analysis (PCA), linear discriminant analysis (LDA), marginal Fisher analysis (MFA), maximum margin criterion (MMC) \cite{ref-book6}, locality preserving projections (LPP) \cite{ref-book7}, sparsity preserving projection (SSP) \cite{ref-journal16}, semi-supervised dimensionality reduction (SSDR), semi-supervised discriminant analysis (SDA) \cite{ref-book8} and random projection (RP) \cite{ref-journal17,ref-journal18}. In particular, random projection is a commonly used technique in data mining and machine learning. It reduces the dimensionality of high-dimensional data by mapping it to a lower dimensional space while preserving the structure of the original data as much as possible. This method is particularly useful when dealing with large-scale datasets, as it can significantly reduce computational and storage requirements. Wright et al.\cite{ref-journal6} examined the applications of RP in the dimension reduction of face images. In the subsequent experiments, we shall also employ random projection technology to conduct dimensionality reduction of the data and realize face recognition.
	
	\section{Experimental results}
	
	In this section, we will conduct experiments based on some public face recognition datasets. These experiments can not only demonstrate the effectiveness of the proposed classification algorithm but also verify the claims made in the previous chapters. Secondly, the robustness of the proposed algorithm against distortion and occlusion shall also be discussed. Particularly, we consider the following four face datasets: the ORL face dataset \cite{ref-journal19}, the Yale face dataset \cite{ref-journal20}, the AR face dataset \cite{ref-journal21} and the FERET face dataset \cite{ref-journal22}. In these experiments, the ORL face dataset contains a total of 400 images of 40 different individuals. The Yale Face Dataset was created by the Yale University's Center for Computational Vision and Control. The dataset comprises 165 images from 15 volunteers, exhibiting variations in lighting conditions, facial expressions, and body poses. The AR dataset comprises over 4,000 frontal images of 126 individuals, with each individual contributing 26 photographs. For this experiment, 10 photos of 40 individuals from this database are selected for recognition. The FERET face dataset contains a total of 1400 face images from 200 individuals, with each person having 7 images. The face images in the dataset include variations in expressions, lighting, and poses. Besides, all the images from the FERET dataset are stored in the TIFF format as grayscale images, with a width of 80 and a height of 80.
	
	Firstly, for the case in the absence of damage and occlusion of the face images, the corresponding recognition rates of various methods are listed in Table \ref{tab1}, where the proposed WWCSC is compared with some existing competitors, such as the SRC, RSRC, Sparse Huber (SH), IRGSC and WHCSC.
	\begin{table}[H]
		\caption{Comparison table of recognition rates for different datasets(Unit:percentage).}
		\centering
		\begin{tabular}{ccccc}
			\toprule
			\textbf{} & \textbf{AR}	& \textbf{ORL} & \textbf{YALE} & \textbf{FERET}\\
			\midrule
			SRC		& 94.17	& 89.17 & 95.56 & 85.36 \\
			RSRC	&95.38	&90.21	&96.34	&88.42\\
			SH	&94.26	&88.32	&93.12	&83.27\\
			IRGSC	&\textbf{96.88}	&\textbf{90.62}	&\textbf{100}	&\textbf{93.35}\\
			WHCSC	&93.21	&89.57	&93.36	&90.42\\
			WCSC	&95.00	&89.00	&94.57	&92.27\\
			WWCSC	&95.00	&90.00	&95.56	&92.78\\
			\bottomrule
		\end{tabular}
		\label{tab1}
	\end{table}
	
It can be inferred from Table \ref{tab1} that the WWCSC method outperforms the WCSC for the four datasets, which implies that adding weights to the loss function can enhance the recognition rate of the model in face recognition. Compared with the SRC, WHCSC and SH methods, the WWCSC method demonstrates superior performance across all four aforementioned datasets, particularly on the FERET dataset. As can be seen from Table \ref{tab1}, on the above four datasets, the performance of IRGSC is the best compared with other methods. This result is attributed to the superiority of the IRGSC method itself. However, the performance of the WWCSC method we proposed is only slightly inferior to that of IRGSC. This result encourages us to explore the robustness of the WWCSC method. Therefore, in the following experiments, we mainly consider the case when there exists loss or occlusion in the face images. In particular, the robustness of the WWCSC is investigated under various types of occlusions, such as the Gaussian noise random pixel corruption and random block occlusion.

\begin{figure}[H]
	\centering
	\includegraphics[width=12 cm]{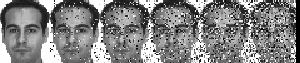}
	\caption{Face images with different percentages of pixel corruption(from 0\% to 50\%).\label{fig1}}
\end{figure}

Secondly, our main research focuses on the robustness of the WWCSC method when face images have different degrees of loss. In this experiment, we specifically conduct our analysis using the AR dataset. For the AR dataset, we artificially damage the face images under varying degrees following Wright,J.\cite{ref-journal6}, where different percentages of randomly chosen pixels from each of the test images are replaced with simulated values from a uniform distribution. For example, effects of some specific face image with various pixel noises are shown in Figure \ref{fig1}.
\unskip

In the followings, in order to validate the robustness of the WWCSC method, we imposed damage ranging from 10\% to 90\% on the test images. The recognition rates of face recognition for different methods with different degrees of damages are presented in Table \ref{tab2}.
	\begin{table}[H]
	\caption{Comparison table of recognition rates of different corruption(Unit:percentage)}
	\centering
	\begin{tabular}{cccccccccc}
		\toprule
	\textbf{} & \textbf{10\%}	& \textbf{20\%} & \textbf{30\%} & \textbf{40\%} & \textbf{50\%} & \textbf{60\%} & \textbf{70\%} & \textbf{80\%} & \textbf{90\%}\\
	\midrule
	SRC	&94.16	&93.33	&93.33	&87.50	&87.50	&84.38	&84.38	&81.25	&65.62 \\
	RSRC	&95.21	&94.37	&93.89	&90.65	&88.47	&85.36	&84.89	&82.37	&70.46\\
	SH	&93.98	&93.41	&92.06	&89.39	&85.72	&83.61	&80.42	&70.15	&60.32\\
	IRGSC	&\textbf{96.88}	&\textbf{96.88}	&94.62	&90.38	&89.25	&89.25	&89.25	&80.12	&72.50\\
	WHCSC	&93.06	&92.58	&91.34	&88.49	&86.67	&84.52	&81.94	&78.61	&66.72\\
	WCSC	&95.00	&95.00	&\textbf{95.00}	&93.36	&92.34	&90.25	&88.23	&86.34	&75.27\\
	WWCSC	&95.00	&95.00	&\textbf{95.00}	&\textbf{94.17}	&\textbf{94.17}	&\textbf{94.17}	&\textbf{94.17}	&\textbf{94.17}	&\textbf{94.17}\\
		\bottomrule
	\end{tabular}
	\label{tab2}
\end{table}

From the Table \ref{tab2}, it can be seen that as the degree of image damage increases, the face recognition rates of most models show a downward trend. Among them, the performance of the WWCSC is relatively stable. When the loss of face images is 10\% or 20\%, the recognition rate of the WWCSC method is marginally lower than those of the IRGSC and RSRC methods. However, when face images suffer from more than 30\% degree of damages, the performance of the WWCSC method surpasses the most of all other methods. In particularly, in the case of the damages of face images reach to 90\%, the recognition rate of the WWCSC method is 28.55\%, 23.71\%, 33.85\%, 21.67\%, 27.45\% and 18.90\% higher than SRC, RSRC, SH, IRGSC, WHCSC and WCSC. Moreover, it can be discerned that the WWCSC is more robust when dealing with damaged images in contrast to some other approaches.  

\begin{figure}[H]
	\centering
	\includegraphics[width=12 cm]{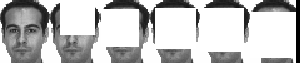}
	\caption{Face images of a white block occlusion(from 50\% to 90\%).\label{fig2}}
\end{figure}

\begin{figure}[H]
	\centering
	\includegraphics[width=12 cm]{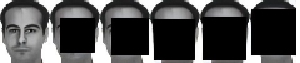}
	\caption{Face images of a black block occlusion(from 50\% to 90\%).\label{fig3}}
\end{figure}

Thirdly, we consider the case when the test images are imposed white or black block occlusions of varying sizes, as shown in Figure \ref{fig2} and Figure \ref{fig3}.


Also, the face recognition rates of different methods under different degrees of damages are listed in Table \ref{tab3}.
	\begin{table}[H]
	\caption{Comparison table of recognition rates of different corruption(Unit:percentage).}
	\centering
	\begin{tabular}{cccccccc}
		\toprule
		\textbf{} & \textbf{10\%}	& \textbf{20\%} & \textbf{30\%} & \textbf{40\%} & \textbf{50\%} & \textbf{60\%} & \textbf{70\%} \\
	\midrule
	SRC	&90.62	&90.62	&84.38	&84.38	&71.88	&65.00	&53.12 \\
	RSRC &91.37	&91.37	&86.43	&82.31	&70.37	&64.35	&55.27\\
	SH	&90.54	&90.54	&85.61	&80.47	&68.35	&60.59	&51.14\\
	IRGSC &\textbf{96.88} &\textbf{93.75}	&87.50	&84.38	&78.12	&68.75	&62.50\\
	WHCSC &92.48 &92.48	&86.87	&83.94	&75.46	&69.68	&60.75\\
	WCSC &94.17	&93.33	&88.33	&87.50	&86.67	&85.83	&79.17\\
	WWCSC &95.00	&93.33	&\textbf{90.83}	&\textbf{90.83}	&\textbf{90.00}	&\textbf{87.50}	&\textbf{80.83}\\
		\bottomrule
	\end{tabular}
	\label{tab3}
\end{table}

 It can be observed from the Table \ref{tab3}  that as the occluded part of the image increases, the face recognition rate of all models shows a gradually decreasing trend. This result indicates that as the proportion of face images that are occluded increases, the difficulty of face recognition also increases. When images exhibit 10\% to 20\% block occlusion, the IRGSC model demonstrates the optimal performance among all evaluated models, with the WWCSC model ranking second in effectiveness. However, when images are subject to 30\% to 70\% block occlusion, recognition rates of the WWCSC distinctly exceed those of  other alternative methods. In particularly, in the case of  block occlusion of face images reach to 70\%, the recognition rate of the WWCSC method is 27.71\%, 25.56\%, 29.69\%, 18.33\%, 20.08\% and 1.66\% higher than SRC, RSRC, SH, IRGSC, WHCSC and WCSC. Moreover, this result demonstrates the superiority of the WWCSC method in the case of high block occlusion in face images.

\section{Conclusions}

In this paper, we propose a new wing constrained sparse coding and weighted wing
constrained sparse coding. Compared with some other methods, the advantage of WWCSC lies in its robustness and effectiveness in dealing with complex issues such as occlusion and damage in images. On the one hand, the wing loss function utilized by the model can excellently diminish the effects of noise or outliers. On the other hand, the model can distinguish face images of different classes by applying weights to the loss function to decrease the intra-class variance and increase the inter-class variance. Experiments also show that WWCSC is superior to IRGSC, RSRC, SRC and so on. The high robustness and strong effectiveness of the WWCSC method demonstrate that it is an ideal choice for face recognition applications.
	

\end{document}